\documentclass{article}
\usepackage{spconf,amsmath,graphicx}

\usepackage{enumitem}
\setlist{nosep, leftmargin=14pt}

\usepackage{mwe} 
\usepackage{dsfont}
\usepackage{subcaption}
\usepackage{easyReview}
\usepackage{hyperref}
\usepackage{multirow}
\usepackage{booktabs}  
\usepackage{siunitx} 
\usepackage{acro}

\DeclareAcronym{kNN}{
short = $k$-NN ,
long = $k$-Nearest Neighbor ,
short-plural = s ,
long-plural = s
}

\DeclareAcronym{MVF}{
short = MVF ,
long = most valuable feature embedding ,
short-plural = s ,
long-plural = s
}

\title{Integrating kNN with Foundation Models for Adaptable and Privacy-Aware Image Classification}
\name{Sebastian Doerrich$^\ast$\thanks{$^\ast$ These authors contributed equally to this work.} \qquad Tobias Archut$^\ast$ \qquad Francesco Di Salvo \qquad Christian Ledig}
\address{xAILab, University of Bamberg, Germany}
\begin{document}
\maketitle
\begin{abstract}
Traditional deep learning models implicity encode knowledge limiting their transparency and ability to adapt to data changes. Yet, this adaptability is vital for addressing user data privacy concerns. We address this limitation by storing embeddings of the underlying training data independently of the model weights, enabling dynamic data modifications without retraining. Specifically, our approach integrates the $k$-Nearest Neighbor ($k$-NN) classifier with a vision-based foundation model, pre-trained self-supervised on natural images, enhancing interpretability and adaptability. We share open-source implementations of a previously unpublished baseline method as well as our performance-improving contributions. Quantitative experiments confirm improved classification across established benchmark datasets and the method’s applicability to distinct medical image classification tasks. Additionally, we assess the method's robustness in continual learning and data removal scenarios. The approach exhibits great promise for bridging the gap between foundation models’ performance and challenges tied to data privacy. The source code is available at \href{https://github.com/TobArc/privacy-aware-image-classification-with-kNN}{github.com/TobArc}.
\end{abstract}
\section{Introduction}
\label{sec:intro}
Deep learning has exhibited significant success in diverse domains, notably in natural language processing \cite{Devlin2019,Brown2020} and image classification \cite{dosovitskiy2021,Radford2021,Caron2021}, driven by the evolution of increasingly sophisticated models. These models, empowered by substantial computational resources, excel in capturing intricate patterns and implicit representations within their parameters \cite{Guu2020}. However, the inherent limitation of tying knowledge exclusively to model weights introduces a significant drawback. The opacity of this knowledge restricts efficient information retrieval \cite{Nakata2022} and raises concerns about data usage rights and privacy \cite{Xu2019}. This challenge is accentuated by evolving regulations, such as the European Union’s \textit{right to erasure (‘right to be forgotten’)} (Article 17 of the General Data Protection Regulation (GDPR) \cite{EUGDPRArticle17}), empowering users to revoke data usage rights promptly.

Updating knowledge in deep learning models, involving tasks such as addition, deletion, or modification of information, currently necessitates comprehensive retraining or fine-tuning \cite{wang2023}. This process incurs substantial computational expenses and proves cumbersome, particularly in sensitive sectors like healthcare. The predominant paradigm of exclusive data storage within model parameters lacks adaptability, especially when users exercise their right to update or delete personal data. This leads to exponential costs and the risk of catastrophic forgetting in continual learning scenarios \cite{Lange2022}, rendering these models unpracticable at best, infeasible, or irresponsible at worst.

In response to these challenges, our research is inspired by Nakata et al.'s solution \cite{Nakata2022}, which deviates from the conventional approach of storing knowledge solely in model parameters. We advocate for storing comprehensive training dataset knowledge, including image feature representations and labels, in an external dynamic repository. This approach enables seamless addition, deletion, or modification of data without necessitating model retraining. Integrating the classical \ac{kNN} classifier \cite{Cunningham2020} with the robust and discriminative feature spaces of foundation models, pre-trained in a self-supervised manner on natural images, enhances interpretability and adaptability. Our contributions encompass:
\begin{itemize}
    \item Open-source implementation including an independent performance validation of Nakata et al.’s work for which there is currently no public implementation available.
    \item Advancing the method’s performance by incorporating recent foundation models and a more flexible data storage system, enabling few-shot adaptation for medical image analysis.
    \item Quantitative confirmation that the method addresses data privacy concerns by facilitating task-incremental learning as well as allowing for data removal in sensitive healthcare scenarios without compromising model performance.
\end{itemize}
%
\section{Related work}
\label{sec:relatedwork}
%
\begin{figure*}
\centering
\includegraphics[width=\linewidth]{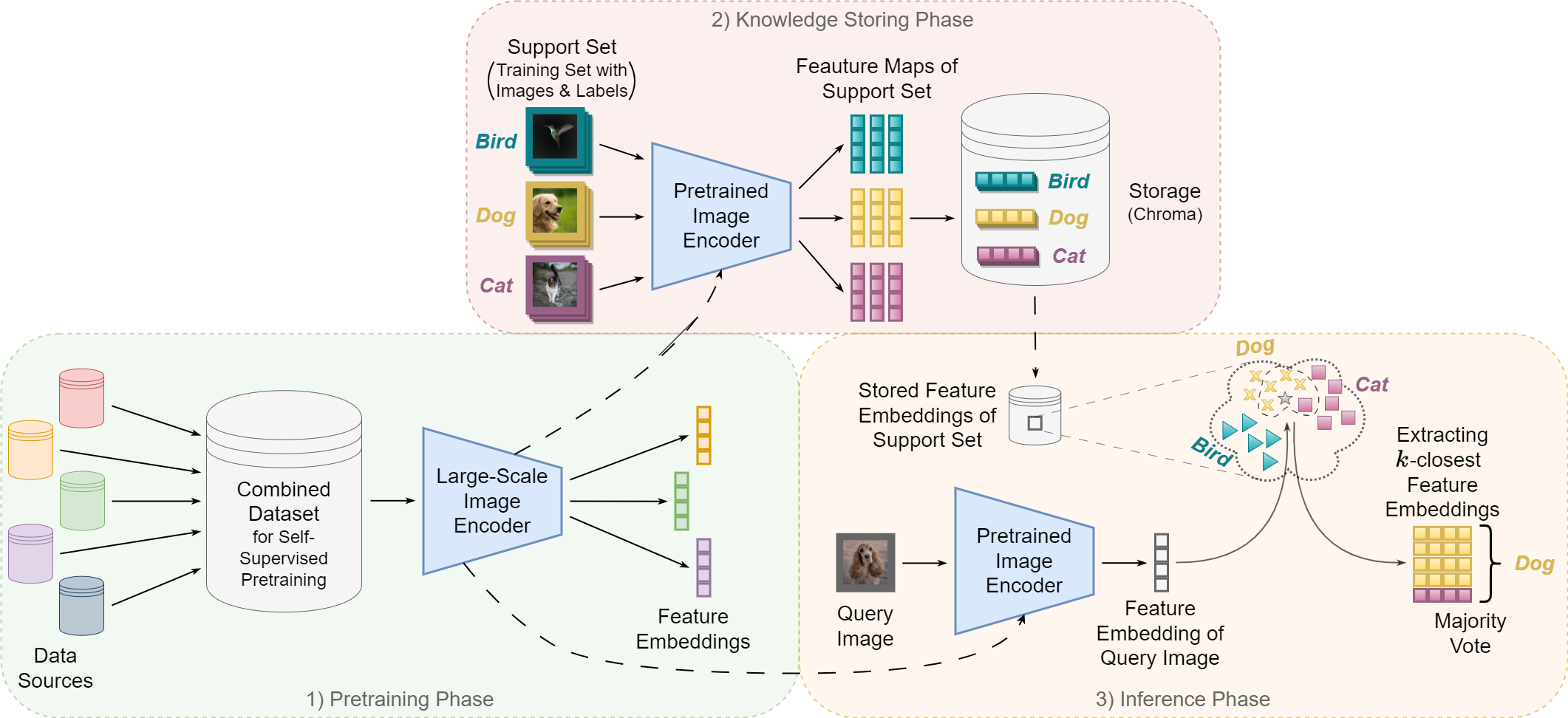}
\caption{During pretraining (1), the image encoder is trained to extract representative features. The knowledge-storing phase (2) utilizes the pre-trained (now frozen) encoder to extract and store task-relevant knowledge from the training data. During inference (3), that knowledge allows the classification of query images through majority voting on the top-$k$ similar embeddings.}
\label{fig:method}
\end{figure*}
Foundation models, exemplified by Transformer \cite{Vaswani2017} and Vision Transformer \cite{dosovitskiy2021}, demand extensive training on large-scale datasets to excel in tasks like natural language processing \cite{Brown2020} or image generation \cite{Ramesh2022}. Self-supervised contrastive methods, such as CLIP \cite{Radford2021} (on image-text pairs) and DINOv2 \cite{oquab2023} (exclusively on image pairs), enable training without intensive labeling, yet fine-tuning on annotated datasets is often necessary to optimize results. However, fine-tuning poses a risk of catastrophic forgetting in continual learning scenarios. Model-based approaches like GEM \cite{LopezPaz2017}, ER \cite{Rolnick2018}, and iCaRL \cite{Rebuffi2016} mitigate this but lock all knowledge in the model's weights, limiting information retrieval and modification possibilities. In contrast, \ac{kNN} methods, previously employed for representation learning evaluation \cite{Caron2021} or noise reduction \cite{bahri2020}, proved promising for enhancing knowledge retrieval as shown by RETRO \cite{Borgeaud2021} in auto-regressive language models.

Nakata et al. \cite{Nakata2022} combine these foundations, employing a \ac{kNN} classifier in a three-phase methodology, which involves pretraining on natural images, knowledge storage through feature map extraction, and inference based on \mbox{\ac{kNN}} retrieval. This eliminates the need for fine-tuning and exhibits efficacy in continual learning scenarios. Our work extends this innovation by integrating the \ac{kNN} classifier with recent vision-based foundation models. Specifically, we propose to extract image features with DINOv2 \cite{oquab2023}, preserving robustness and adaptability across diverse scenarios while enhancing classification. Additionally, separating computation from storage ensures flexible knowledge management, addressing data privacy concerns in particular. We further extend the method and quantitatively confirm its applicability to the sensitive healthcare domain by demonstrating uncompromised performance in challenging task-incremental learning and seamless data removal scenarios.
\section{Method}
\label{sec:method}
\subsection{Overview}
\label{subsec:overview}
Our approach utilizes a three-phase structure as depicted in \figurename~\ref{fig:method} and further outlined below:

\noindent\textbf{Pretraining Phase}\:\:\: Initially, a foundation model is pre-trained on a large-scale dataset, accommodating unlabeled or noisily labeled images to obviate upfront labeling costs. The focus is on extracting generic, vision task-independent features crucial for robust and reliable \ac{kNN} performance. The choice of an image encoder trained on a diverse dataset becomes imperative for effective segregation of feature embeddings, facilitating robust generalization across datasets.
    
\noindent\textbf{Knowledge-Storing Phase}\:\:\: In the knowledge-storing phase, the pre-trained image encoder captures feature embeddings from the training set (support set) which are subsequently stored along with the corresponding labels in an external database. This way, task-relevant knowledge is kept separate from the encoder's weights, adhering to continual learning paradigms and privacy regulations. This design allows seamless addition, modification, and deletion of samples.

\noindent\textbf{Inference Phase}\:\:\: During inference, the pre-trained image encoder generates a feature embedding for a given query image. Top-$k$ similar feature embeddings are retrieved from the external database using cosine similarity as the distance metric. We use cosine similarity due to its robustness in capturing scale-invariant angular relationships between vectors, making it particularly effective for measuring similarity in multi-dimensional data representations \cite{Mulekar2014}. Classification of the query image is determined through a majority vote on the labels associated with the top-$k$ similar feature embeddings, enabling efficient classification without encoder retraining.
\subsection{Backbone architecture}
\label{subsec:backbone}
Differing from Nakata et al. \cite{Nakata2022}, we opt for the DINOv2 \cite{oquab2023} backbone over CLIP \cite{Radford2021} to enhance the robustness of our method. DINOv2 employs self-supervised contrastive training on 142 million distinct images from curated and uncurated data sources, emphasizing high-quality feature representation by minimizing the distance of similar objects and maximizing the distance of distinct ones. We choose \mbox{DINOv2} Large\footnote{\texttt{vit\_large\_patch14\_dinov2.lvd142m}}\addtocounter{footnote}{+1}\addtocounter{Hfootnote}{+1}\footnotemark\addtocounter{footnote}{-1}\addtocounter{Hfootnote}{-1}\addtocounter{footnote}{-1}\addtocounter{Hfootnote}{-1} with $14 \times 14$ patches and $1024$-sized image embeddings over its Base\footnote{\texttt{vit\_base\_patch14\_dinov2.lvd142m}}\footnote{\label{HuggingFace}\texttt{\url{https://huggingface.co/timm}}} version, to increases model capacity and feature representation (304M parameters vs. 87M parameters).
\subsection{Knowledge storage}
\label{subsec:storage}
Nakata et al. \cite{Nakata2022} require loading both the image encoder model and all stored feature embeddings into a singular processing unit's memory, resulting in significant computational demands, especially for large support sets. This imposes a continuous need for substantial computational resources. To mitigate this challenge, our strategy involves the active separation of storage from computational processes. Utilizing Chroma \cite{Huber2023}, an open-source, in-memory embedding database, we ensure efficient storage and retrieval of feature embeddings. It's essential to note that alternative vector database solutions could be employed including highly efficient approximate nearest neighbor search algorithms.
\section{Experiments and Results}
\label{sec:experiments}
We first evaluate the choice of backbone in terms of our method's classification ability on natural images. We further assess the adaptability of our approach to image classification tasks in the medical domain including its ability for task-incremental learning and its potential for seamless removal of sensitive, task-relevant data without seriously compromising performance. To this end, we utilize a comprehensive set of distinct datasets comprising natural images, such as CIFAR-10 \cite{Krizhevsky2012}, CIFAR-100 \cite{Krizhevsky2012}, and STL-10 \cite{Coates2011}, as well as two datasets comprising medical images, namely the Pneumonia Dataset \cite{KERMANY2018}, depicting pediatric chest X-ray images of patients with and without pneumonia, and the Melanoma Skin Cancer Dataset of the 2018 ISIC
challenge \cite{codella2019,Tschandl2018}, depicting benign and different malignant melanoma images. Further details are described in \tablename~\ref{tab:dataset}. To allow a fair comparison with Nakata et al.'s method, we employ the same $k$ ($k = 10$) for the \ac{kNN} classifier throughout all our experiments. 
\begin{table}
\caption{Details of the selected datasets ($^\ast$ the reported resolution represents the average resolution across all samples).}
\label{tab:dataset}
    \centering
    \begin{tabular}{l c c c c }
        \toprule
        Dataset & \# C & Resolution & \# Train / Test \\
        \midrule
        CIFAR-10 & $10$ & $3 \times 32 \times 32$ & $50,000$ / $10,000$ \\
        CIFAR-100 & $100$ & $3 \times 32 \times 32$ & $50,000$ / $10,000$ \\
        STL-10 & $10$ & $3 \times 96 \times 96$ & $5,0000$ / $8,000$ \\
        Pneumonia$^\ast$ & $2$ & $1 \times 1328 \times 971$ & $5,232$ / $624$ \\
        Melanoma & $7$ & $3 \times 600 \times 450$ & $10,015$ / $1,513$ \\
        \bottomrule
    \end{tabular}
\end{table}
%
\subsection{Backbone choice for our method}
\label{subsec:expbackbone}
To validate our backbone choice, we compare the classification performance of our method with DINOv2 Large, to DINOv2 Base, a WideResNet101 \cite{zagoruyko2017} pre-trained on ImageNet-1k \cite{Deng2009} as well as our own implementation of Nakata et al.'s ViT-B/16\addtocounter{Hfootnote}{-1}\addtocounter{footnote}{-1}\footnotemark\footnote{\texttt{vit\_base\_patch16\_clip\_224.openai}} and ViT-L/14\addtocounter{Hfootnote}{-1}\addtocounter{footnote}{-1}\addtocounter{Hfootnote}{-1}\addtocounter{footnote}{-1}\footnotemark\addtocounter{Hfootnote}{+1}\addtocounter{footnote}{+1}\footnote{\texttt{vit\_large\_patch14\_clip\_336.openai}} image encoder models (both pre-trained by CLIP). The results on CIFAR-10, CIFAR-100, and STL-10 are presented in \tablename~\ref{tab:classification}. The results demonstrate the overall increased representative ability of models pre-trained in a self-supervised fashion compared to supervised pretraining. Moreover, both DINOv2 models, in particular DINOv2 ViT-L/14, showcase superior classification prowess compared to CLIP, endorsing the benefits of embracing self-supervised, image-exclusive pre-training for image-specific tasks.
%
\begin{table}
\caption{Classification accuracy of our \ac{kNN} approach for different backbone choices.}
\label{tab:classification}
    \centering
    \begin{tabular}{l c c c}
        \toprule
        Accuracy [\%] & CIFAR-10 & CIFAR-100 & STL-10 \\
        \midrule
        $\text{ResNet-101}$ & $87.3$ & $63.6$ & $98.1$ \\
        $\text{CLIP ViT-B/16}$ & $92.4$ & $68.0$ & $98.5$ \\
        $\text{CLIP ViT-L/14}$ & $95.5$ & $74.2$ & $99.4$ \\
        $\text{DINOv2 ViT-B/14}$ & $98.0$ & $87.2$ & $99.4$ \\
        $\text{DINOv2 ViT-L/14}$ & $\mathbf{98.5}$ & $\mathbf{88.3}$ & $\mathbf{99.5}$ \\
        \bottomrule
    \end{tabular}
\end{table}
\subsection{Adaptation for medical image analysis}
\label{subsec:medicaldata}
To evaluate the applicability of our \ac{kNN} method in the medical domain, we first assess its classification performance on the Pneumonia and Melanoma dataset. We compare the performance of our approach with state-of-the-art, fully supervised benchmarks trained end-to-end. For Pneumonia, we compare to CovXNet by Mahmud et al. \cite{MAHMUD2020103869} and for Melanoma to Cassidy et al.'s EfficientNetB0 model \cite{CASSIDY2022102305}. The results are displayed in \tablename~\ref{tab:medicalclassification}. Despite the distinct, transferred behavior of this task, DINOv2 does not employ any medical knowledge during training, our method demonstrates high classification potential, even surpassing the supervised state of the art for the Melanoma dataset.
\begin{table}
\caption{Comparison of our approach's strong transfer learning ability for medical image analysis. ($\vphantom{x}^\dag$ refers to fully supervised models, trained end-to-end.)}
\label{tab:medicalclassification}
    \centering
    \begin{tabular}{l c c}
        \toprule
        Accuracy [\%] & Pneumonia & Melanoma \\
        \midrule
        $\text{CovXNet}^{\dag}$ \cite{MAHMUD2020103869} & $\mathbf{98.1}$ & --- \\
        $\text{EfficientNetB0}^{\dag}$ \cite{CASSIDY2022102305} & --- & $62.1$ \\
        $\text{Ours (DINOv2 ViT-B/14)}$ & $88.1$ & $68.5$ \\
        $\text{Ours (DINOv2 ViT-L/14)}$ & $89.9$ & $\mathbf{69.8}$ \\
        \bottomrule
    \end{tabular}
\end{table}
\subsection{Continual learning and incremental forgetting}
\label{subsec:continuallearning}
\begin{figure}
\centering
\begin{minipage}[b]{0.48\linewidth}
  \centering
  \centerline{\includegraphics[width=\linewidth]{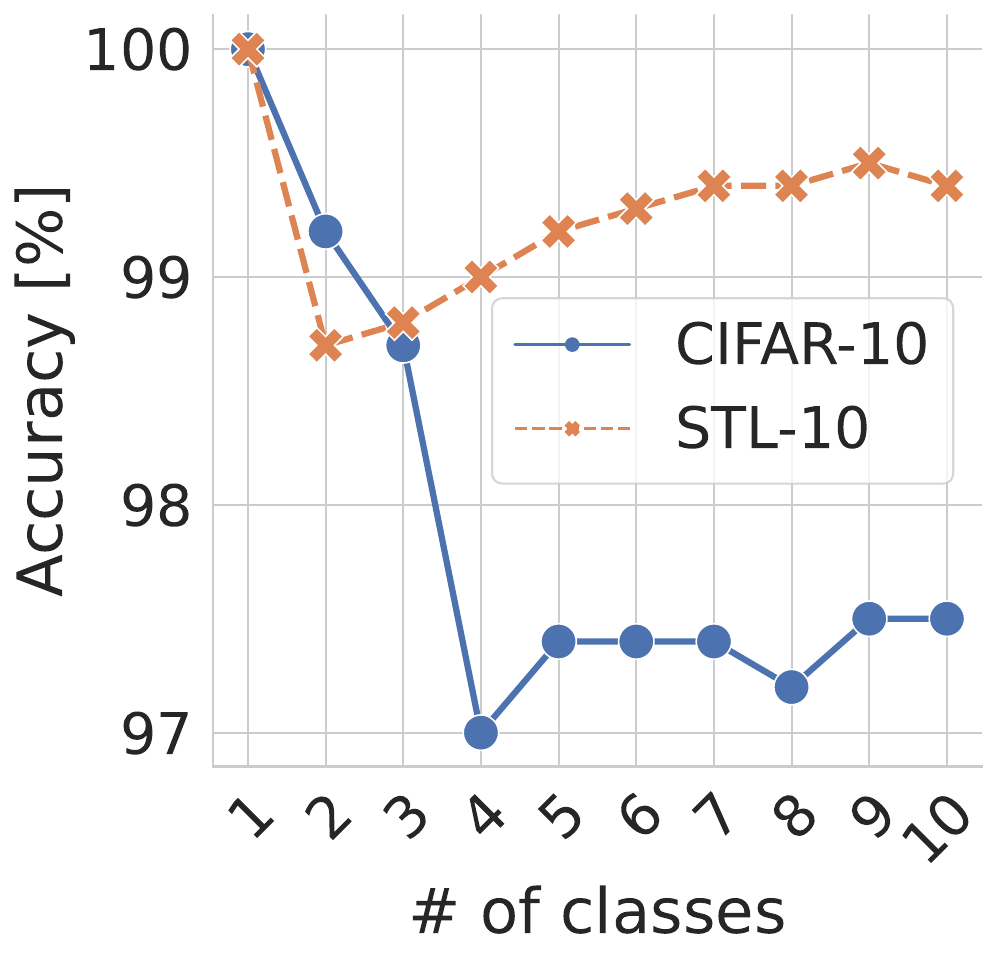}}
  \centerline{(a) Class incremental learning}\medskip
\end{minipage}
\hfill
\begin{minipage}[b]{0.48\linewidth}
  \centering
  \centerline{\includegraphics[width=\linewidth]{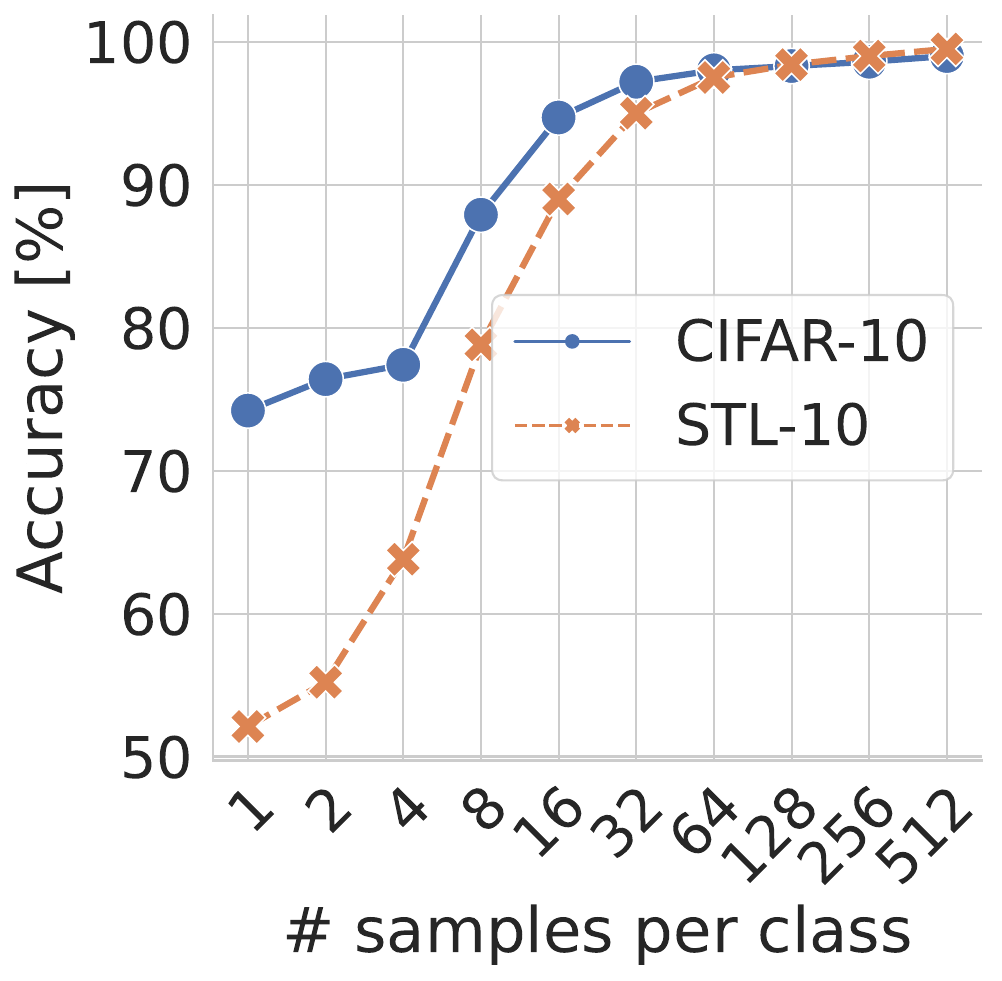}}
  \centerline{(b) Sample incremental learning}\medskip
\end{minipage}
\caption{Visualization of the method's ability for diverse continual learning tasks.}
\label{fig:continuallearning}
\end{figure}

Nakata et al. \cite{Nakata2022} have shown that the \ac{kNN} approach promises the potential to mitigate catastrophic forgetting in continual learning scenarios for natural image datasets when incrementally adding additional classes or samples of existing classes to the support set. We first confirm this potential on CIFAR-10 and STL-10, by incrementally adding entirely new classes to the support and test set, as well as incrementally adding additional feature embeddings to the support set and evaluating the classification performance. \figurename~\ref{fig:continuallearning} (a) and \figurename~\ref{fig:continuallearning} (b) present the results for each task, respectively, showcasing the constant classification performance of our method for the class incremental learning task as well as a remarkable classification ability for the sample incremental learning task already for only a few samples per class in the support set. By using an adaptive $k$ instead of our fixed $k$, this few-shot classification capability could be improved even further.

Additionally, we evaluate the incremental learning capability of our method when transferring it to the medical domain. However, this time, we assess our method for incrementally adding datasets of different anatomies and distributions, instead of sticking to the same domain by adding additional classes of the same dataset. For this, we compare the method's exclusive performance on the Pneumonia and Melanoma dataset (cf. \tablename~\ref{tab:medicalclassification}) with its performance on a combined version of both datasets, which comprises a diverse distribution in a multi-class classification setting. Notably, the accuracy on the exclusive datasets is nearly consistent with the accuracy on the combined version (\qty{89.9}{\%} vs. \qty{89.9}{\%} for Pneumonia and \qty{69.8}{\%} vs. \qty{69.0}{\%} for Melanoma).

\begin{figure}
\centering
\begin{minipage}[b]{0.48\linewidth}
  \centering
  \centerline{\includegraphics[width=\linewidth]{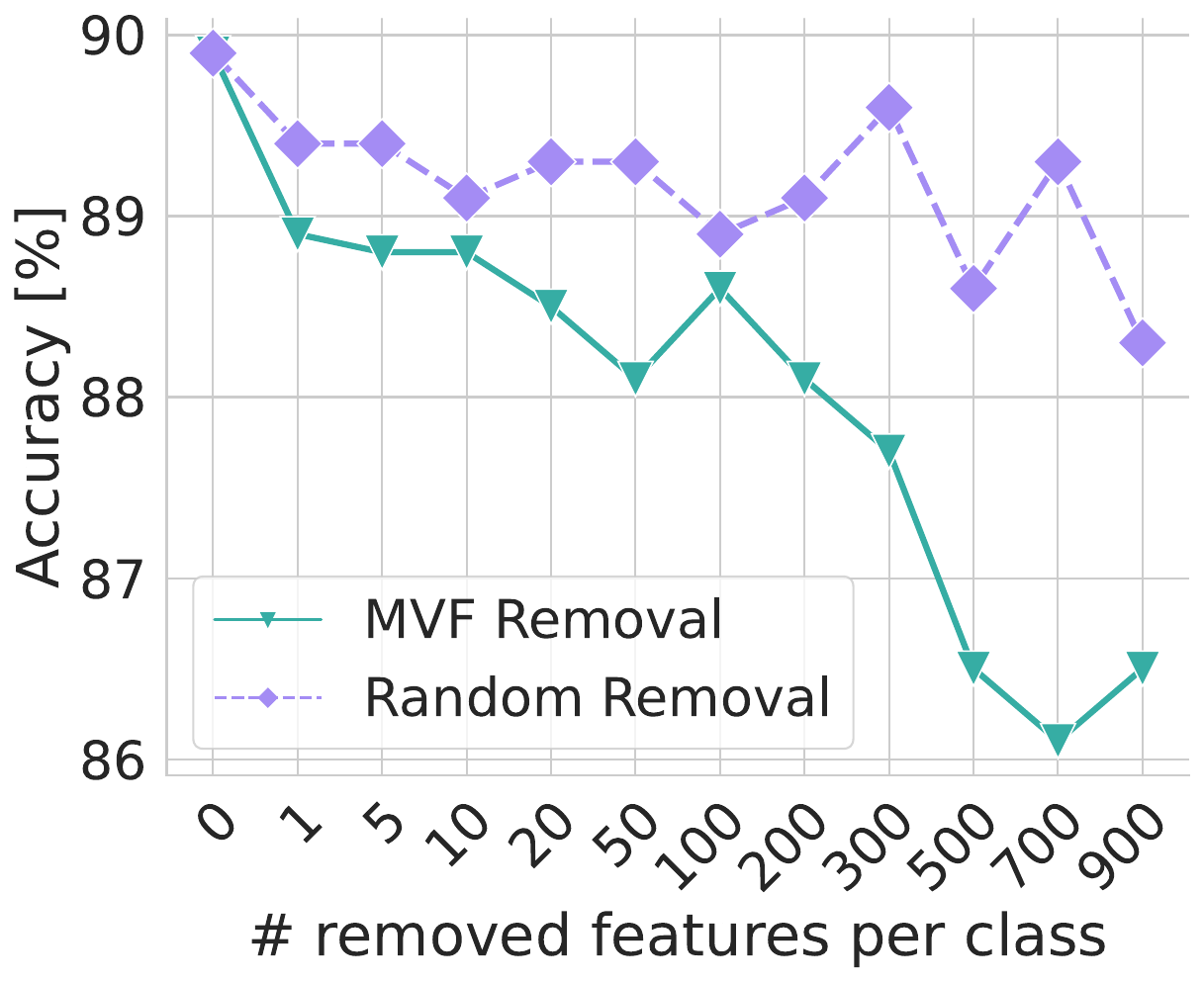}}
  \centerline{(a) Pneumonia}\medskip
\end{minipage}
\hfill
\begin{minipage}[b]{0.48\linewidth}
  \centering
  \centerline{\includegraphics[width=\linewidth]{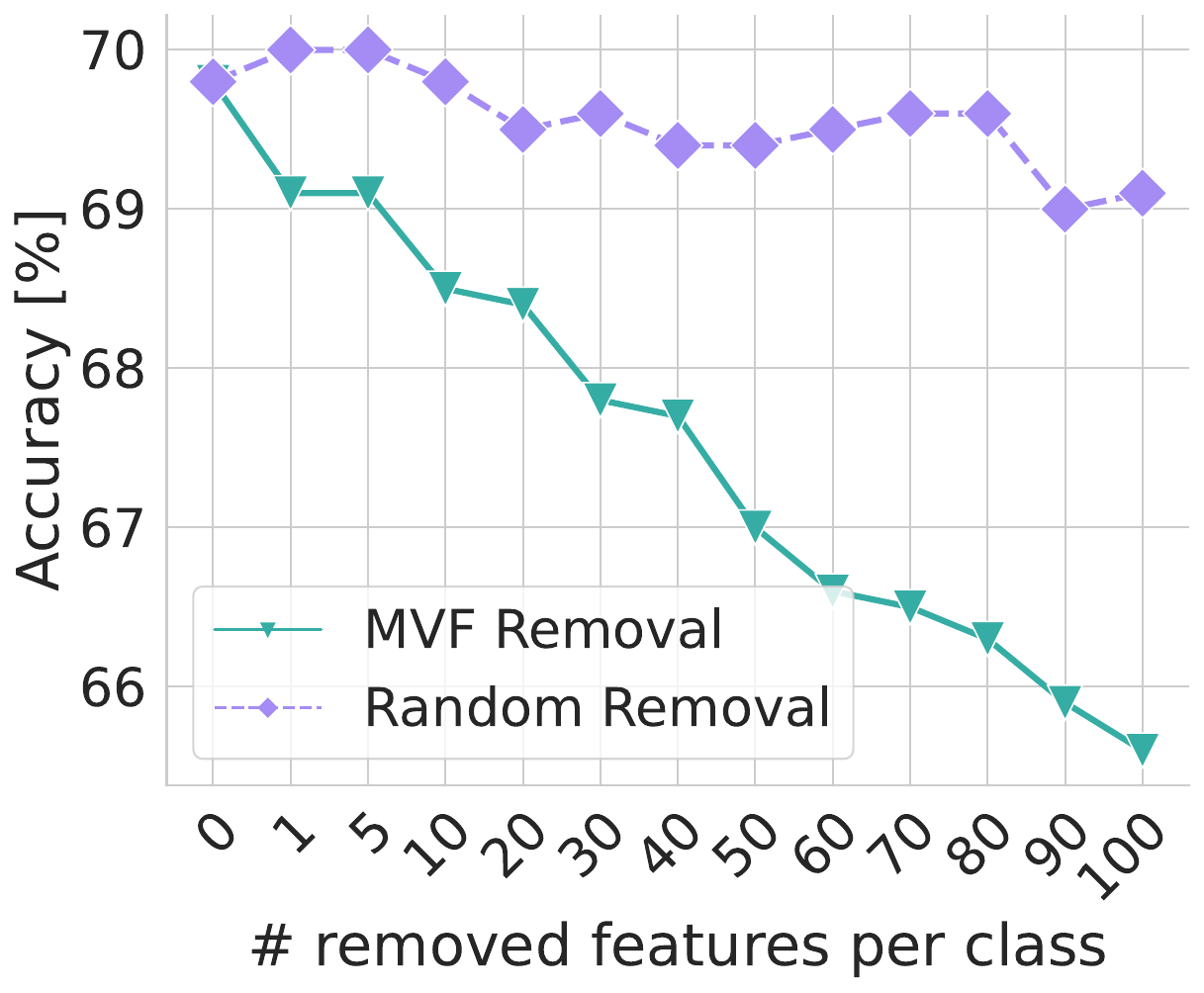}}
  \centerline{(b) Melanoma}\medskip
\end{minipage}
\caption{Illustration of our method's classification consistency despite the continuous diminishing of the support set.}
\label{fig:incrementalforgetting}
\end{figure}

Lastly, we investigate our approach's ability to facilitate the effortless removal of task-relevant data, ensuring minimal impact on the model's performance, particularly when patients exercise their rights to revoke data usage and demand the deletion of their information from the model. To this end, we evaluate the impact on classification performance if we remove either random samples or if we remove the \ac{MVF} of each class from the support set. One class’s \ac{MVF} is that feature embedding of the support set which the \ac{kNN} algorithm utilizes the most to correctly classify query samples during inference on a fixed test set. In other words, this feature embedding contributes the most to the classification performance of the method. \figurename~\ref{fig:incrementalforgetting} visualizes this for the Pneumonia and the Melanoma dataset. The results present that removing nearly all support set samples poses only a slight, negative impact on the overall classification performance, demonstrating the few-shot ability of our model once again and thus demonstrating the overall potential of our method to remove any knowledge from our model without severely impairing the classification performance.
\section{Discussion and Conclusion}
\label{sec:conclusion}
In this work, we present an open-source, improved version of the \ac{kNN} integration with vision-based foundation models, originally proposed by Nakata et al. \cite{Nakata2022} that was not made publicly available by the authors before. Extensive experiments present our method's classification ability, apparent due to its high classification accuracy on natural images. We affirm its suitability for continuous learning scenarios, preventing catastrophic forgetting. Moreover, we showcase its potential for application in the medical domain, owing to its robust out-of-the-box performance and ability to seamlessly remove task-relevant data with minimal impact on performance. Our approach represents a significant step towards bridging the gap between foundation models' great performances and the challenges of data accessibility, privacy, and adaptability.
\section{Compliance with ethical standards}
\label{sec:ethics}
This research study was conducted retrospectively using human subject data made available in open access. Ethical approval was not required as confirmed by the license attached with the open-access data.
%
\bibliographystyle{IEEEbib}
\bibliography{refs_extensive}

\end{document}